\documentclass{article}
\usepackage{spconf}
\usepackage{amsmath,graphicx}

\usepackage{enumitem}
\usepackage{amsthm}
\usepackage{amsmath,amsfonts,amssymb,url} 
\usepackage{adjustbox}
\usepackage{graphicx}
\usepackage{hyperref}
\usepackage{setspace}
\usepackage{epsfig}  
\usepackage{url}  
\usepackage{lscape}  
\usepackage[justification=raggedright]{caption}	
\usepackage{multirow}
\usepackage{multicol}
\usepackage{float,graphicx}
\usepackage{subcaption}
\usepackage[english]{babel}
\usepackage{bbm}
\usepackage{prettyref,xspace}
\usepackage{cite}
\usepackage{multirow}

\newrefformat{eq}{(\ref{#1})}
\newrefformat{thm}{Theorem~\ref{#1}}
\newrefformat{th}{Theorem~\ref{#1}}
\newrefformat{chap}{Chapter~\ref{#1}}
\newrefformat{sec}{Section~\ref{#1}}
\newrefformat{algo}{Algorithm~\ref{#1}}
\newrefformat{fig}{Figure.~\ref{#1}}
\newrefformat{tab}{Table~\ref{#1}}
\newrefformat{rmk}{Remark~\ref{#1}}
\newrefformat{clm}{Claim~\ref{#1}}
\newrefformat{def}{Definition~\ref{#1}}
\newrefformat{cor}{Corollary~\ref{#1}}
\newrefformat{lmm}{Lemma~\ref{#1}}
\newrefformat{prop}{Proposition~\ref{#1}}
\newrefformat{pr}{Proposition~\ref{#1}}
\newrefformat{app}{Appendix~\ref{#1}}
\newrefformat{ques}{Question~\ref{#1}}

\theoremstyle{definition}

\usepackage{amsmath}
\usepackage{amssymb}
\usepackage{latexsym}
\usepackage{verbatim}
\usepackage{psfrag}
\usepackage{epsfig}
\usepackage{xcolor}
\usepackage[ruled,vlined]{algorithm2e}

\SetKwInput{KwInput}{Input}                
\SetKwInput{KwOutput}{Output} 
\SetKwInput{KwParameters}{Parameters} 
\newcommand{\Ma}[1]{\ensuremath{\mathbf{#1}}}
\newcommand{\Ve}[1]{\ensuremath{\mathbf{#1}}}

\newcommand{\Real}{\ensuremath{\mathbb{R}}}

\newcommand{\R}{\Real}

\newcommand{\norm}[1]{\left\Vert#1\right\Vert}

\newcommand{\mI}{\Ma{I}}

\newcommand{\mL}{\Ma{L}}

\newcommand{\vu}{\Ve{u}}
\newcommand{\vx}{\Ve{x}}
\newcommand{\vy}{\Ve{y}}

\newcommand{\calN}{\ensuremath{\mathcal{N}}}

\newcommand{\bb}{\begin{bmatrix}}
\newcommand{\eb}{\end{bmatrix}}

\newcommand{\mD}{\Ma{D}}

\newcommand{\mU}{\Ma{U}}
\newcommand{\mV}{\Ma{V}}

\newcommand{\mZ}{\Ma{Z}}
\newcommand{\mX}{\Ma{X}}
\newcommand{\mY}{\Ma{Y}}

\newcommand{\mSigma}{\Ma{\Sigma}}

\newcommand{\vv}{\Ve{v}}

\setlist{nosep, leftmargin=14pt}
\title{Multimodal Fusion using Sparse CCA \\for Breast Cancer Survival Prediction}
\name{Vaishnavi Subramanian$^1$, Tanveer Syeda-Mahmood$^2$, Minh N. Do$^{1,3}$}
\address{$^1$ Electrical and Computer Engineering, University of Illinois at Urbana-Champaign, USA\\
$^2$ IBM Research, Almaden Research Center, San Jose, USA\\
$^3$ College of Engineering and Computer Science, VinUniversity, Hanoi, Vietnam
}
\begin{document}
\maketitle
\let\oldbibliography\thebibliography
\renewcommand{\thebibliography}[1]{%
  \oldbibliography{#1}%
  \setlength{\itemsep}{0pt}%
}
\begin{abstract}
Effective understanding of a disease such as cancer requires fusing multiple sources of information captured across physical scales by multimodal data. In this work, we propose a novel feature embedding module that derives from canonical correlation analyses to account for intra-modality and inter-modality correlations.
Experiments on simulated and real data demonstrate how our proposed module can learn well-correlated multi-dimensional embeddings. These embeddings perform competitively on one-year survival classification of TCGA-BRCA breast cancer patients, yielding average F1 scores up to 58.69\% under 5-fold cross-validation.
\end{abstract}
\section{Introduction}
In a complex disease such as cancer, the interactions between the tumor and host can exist at the molecular, cellular, tissue, and organism levels. Thus, evidence for the disease and its evolution may be present in multiple modalities across scale such as clinical, genomic, molecular, pathological, and radiological imaging. An improved disease understanding requires bridging scales of observation through
multimodal fusion and is the focus of recent cancer research
in survival prediction~\cite{cheerla2019deep,silva2020pan,subramanian2020multimodal} and disease understanding~\cite{subramanian2018correlating,subramanian2018integration}. 
The majority of multimodal techniques focus on representation, alignment, and fusion
of modalities. These include deep learning methods~\cite{silva2020pan,cheerla2019deep,subramanian2020multimodal} and linear analyses
~\cite{subramanian2018correlating,subramanian2018integration,hotelling1936relations}. 

Our focus in multimodal fusion is to identify the shared (common) information present across modalities for obtaining a robust characterization of the underlying problem. Accurate quantification of the shared information should account for the correlations within and across modalities that capture the underlying dependencies.
Of the different fusion techniques, the classical formulation of canonical correlation analysis (CCA)~\cite{hotelling1936relations} has been useful in discovering cross-modality correlations by identifying highly-correlated features from two modalities as a set of canonical variates. 
When applied to cancer, we have earlier shown that CCA enables the discovery of genotype-phenotype associations~\cite{subramanian2018correlating,subramanian2018integration}. 
Although the classical CCA formulation can model the correlations across modalities, it does not explicitly capture the correlations and dependencies of features within each modality. To account for these intra-modality correlations,
group and graph structures can be incorporated~\cite{chen2012efficient,du2015gn} within a sparse CCA framework (SCCA)~\cite{witten2009penalized} to yield higher real data correlations.
Further, while CCA-based feature fusion has been applied for downstream prediction tasks in image recognition~\cite{sun2005new} and autism~\cite{zhao2017feature}, the utility of the learnt canonical variates is yet to be investigated under cancer settings and with sparse variants of CCA.

\begin{figure}[t!]
    \centering
    \includegraphics[trim=20 10 70 10,clip,width=0.48\textwidth]{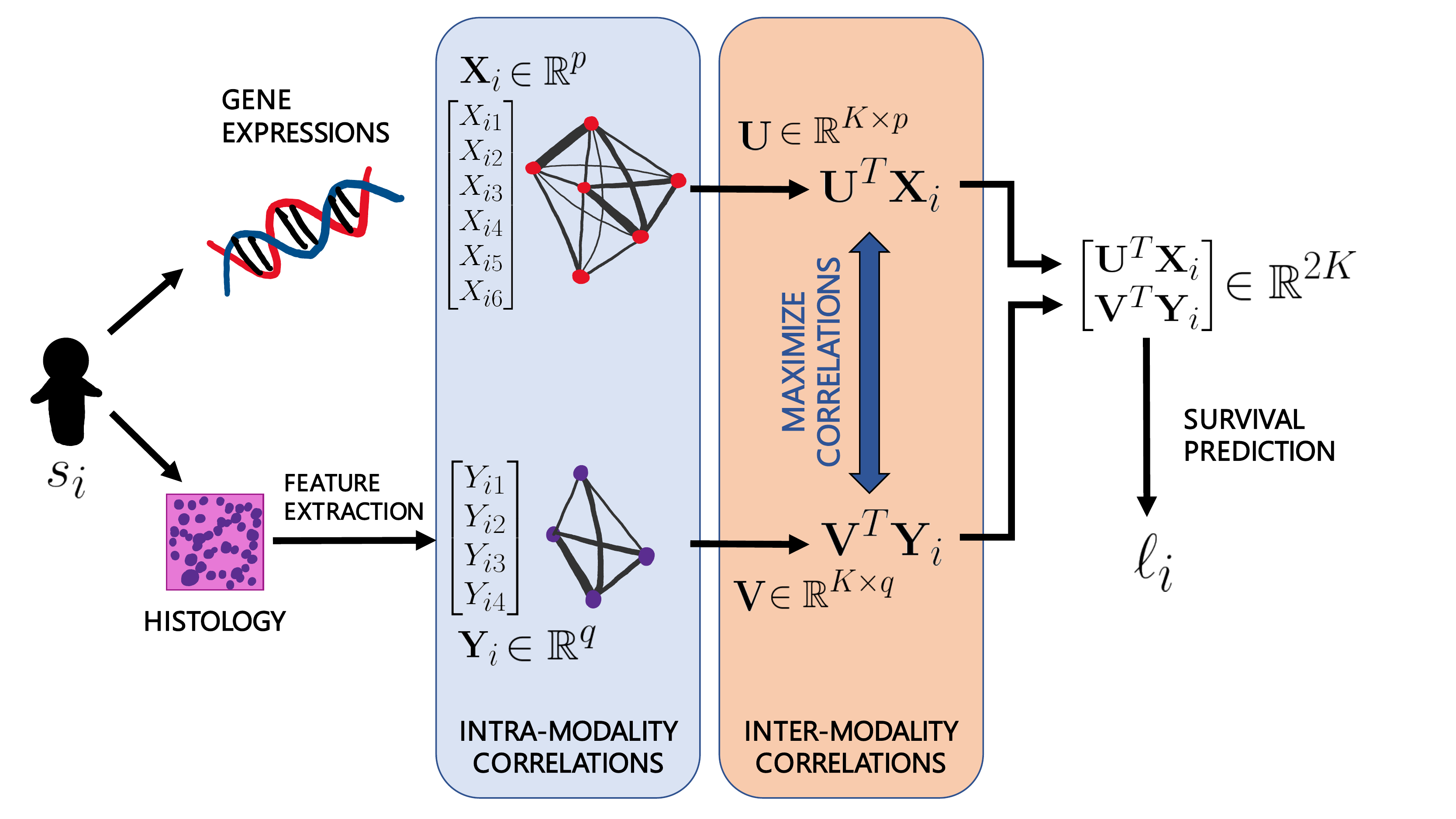}
    \caption{Overview: We make use of intra-modality and inter-modality correlations to learn embedding matrices $\mU$ and $\mV$ which project to well-correlated spaces. The projections are used for downstream prediction tasks. }
    \label{fig:overview}
\end{figure}

In this work, we capture intra-modality correlations through modality-specific graph representations and inter-modality correlations through the CCA objective to develop a linear feature embedding module. We propose an iterative scheme to generate projections of the two modalities' features onto multi-dimensional well-correlated spaces which can be used for downstream prediction tasks such as cancer survival, as outlined in Fig.~\ref{fig:overview}.

\section{Method}
\label{sec:proposed_method}

In this section, we review a graph-structured variant of sparse CCA and present how to generate multi-dimensional embeddings using an iterative approach. Finally, we present our proposed two-modality feature embedding. 
\subsection{Graph-based SCCA (1-GCCA)}
\label{subsection:GN-SCCA}

The CCA formulation~\cite{hotelling1936relations}
considers matched observations of $n$ samples $\mX \in \R^{p \times n}$ and $\mY \in \R^{q \times n}$ from two views. 
The goal of CCA is to identify embedding directions $\vu \in \R^p$ and $\vv \in \R^q$ to maximize the correlation coefficient, between resulting $1$-dimensional embeddings $\vu^T \mX$ and $\vv^T \mY$:
\begin{equation*}
\rho^* =\max_{\vu,\vv} \
\vu^T\mX \mY^T \vv,  \
\text{s.t.}  \  \norm{ \mX^T \vu }_2 = \norm{ \mY^T \vv}_2 =1.
\end{equation*}
Sparse CCA (SCCA)~\cite{witten2009penalized} further imposes sparsity in the entries of the embedding directions $\vu$ and $\vv$ as additional constraints. 
The different features of $\mX$ and $\mY$ often demonstrate intra-modality correlations/dependencies which can be estimated using sample covariance matrices and represented as underlying graphs for each modality.
It is then additionally desired that the entries of embedding directions $\vu$ and $\vv$ weigh well-connected features on the graph similarly, such that underlying properties captured by these well-connected features are highlighted. This is enforced in the 1-dimensional graph-based CCA (1-GCCA) formulation:
\begin{align*} 
\nonumber \max_{\vu, \vv} \ \vu^T \mX \mY^T \vv  \text{ s.t. } & \norm{\vu^T \mX}_2 \leq 1, \norm{\vu}_1 \leq c_1, \vu^T \mL_1 \vu \leq c_0, \\
   & \norm{\vv^T \mY}_2 \leq 1, \norm{\vv}_1 \leq d_1, \vv^T \mL_2 \vv \leq d_0,
\end{align*}
where $c_0, c_1, d_0, d_1$ are constants, $\mL_1$ and $\mL_2$ are the graph Laplacian matrices corresponding to the two respective modalities' underlying graphs. 

This bi-convex problem and can be solved to a local optimum using alternate optimization as shown in Algorithm~\ref{algo:gscca-1}. 
Algorithm~\ref{algo:gscca-1} takes as input the correlation matrices $\mathbf{\Sigma}_x = \mX \mX^T \in \R^{p \times p},\mathbf{\Sigma}_y = \mY \mY^T \in \R^{q \times q},$ cross-correlation matrix $\mathbf{\Sigma}_{xy} = \mX \mY^T  \in \R^{p \times q},$ and graph Laplacians $\mL_1 \in \R^{p \times p}$ and $\mL_2 \in \R^{q \times q}$, and returns embedding vectors $\vu$ and $\vv$. 

\subsection{Multi-dimensional Embeddings (K-GCCA)}

The 1-GCCA algorithm provides a way to identify embedding directions $\vu$ and $\vv$ which maximize the correlations of the 1-dimensional embeddings $\vu^T \mX$ and $\vv^T \mY$. However, 1-dimensional embeddings are often too restrictive and result in information loss. 
We extend the idea of 1-GCCA to identify $K$ embedding directions $\mU=\{\vu^1 \dots \vu^K\} \in \R^{p\times K}$, $\mV=\{\vv^1 \dots \vv^K\} \in \R^{q\times K}$ using the Hotelling's deflation scheme, similar to $K$-factor penalized matrix decomposition~\cite{witten2009penalized}. We obtain 1-dimensional embeddings repeatedly, subtracting the contribution of already found embedding directions using projections followed by normalization.
The algorithm for multi-dimensional graph-based CCA ($K$-GCCA) is described in Algorithm~\ref{algo:gscca-k}.

\begin{algorithm}[t!]
\setstretch{1.15}
\SetAlgoLined
\KwInput{Correlation matrices $\mathbf{\Sigma}_x, \mathbf{\Sigma}_y ,  \mathbf{\Sigma}_{xy}$, graph Laplacian matrices $\mL_1, \mL_2$}
\KwParameters{$\Theta = (\alpha_1, \beta_1, \lambda_1, \alpha_2, \beta_2, \lambda_2) $} 
\KwOutput{Embedding directions $\vu$ and $\vv$}
Initialize $\vu_{(0)} = \mathbf{1}_p/p, \vv_{(0)} = \mathbf{1}_q/q, i=1$ \\
\While{not converged}{
$\mD_u \leftarrow \text{diag}({1}/{|{\vu}_{(i-1)}|}) $,  \
$\mD_v \leftarrow \text{diag}({1}/{|{\vv}_{(i-1)}|}) $\\
${\vu}_{(i)} \leftarrow   (\alpha_1 \mathbf{\Sigma}_x   + \beta_1 \mD_u  +  \lambda_1 \mL_1)^{-1} \mathbf{\Sigma}_{xy} \vv_{(i-1)}$ \\
${\vv}_{(i)} \leftarrow   (\alpha_2  \mathbf{\Sigma}_y + \beta_2 \mD_v  +  \lambda_2 \mL_2)^{-1} \mathbf{\Sigma}_{xy}^T \vu_{(i)}$ 
$i \leftarrow i + 1$
} 
$\vu \leftarrow \dfrac{\vu_{(i-1)}}{\vu_{(i-1)}^T \mSigma_{x} \vu_{(i-1)}}$, $\vv \leftarrow \dfrac{\vv_{(i-1)}}{\vv_{(i-1)}^T \mSigma_y \vv_{(i-1)}}$.
\caption{{$1$-Graph-based SCCA (1-$GCCA$)}}
\label{algo:gscca-1}
\end{algorithm}
\begin{algorithm}[t!]
\setstretch{1.15}
\SetAlgoLined
\KwInput{Data matrices $\mX \in \R^{p \times n}$, $\mY \in \R^{q \times n}$, graph Laplacian matrices
         $\mL_1$, $\mL_2$}
\KwParameters{$K, \Theta = (\alpha_1, \beta_1, \lambda_1, \alpha_2, \beta_2, \lambda_2)$} 
\KwOutput{Embdedding direction matrices $\mU=\{\vu^1 \dots \vu^K\}$, $\mV=\{\vv^1 \dots \vv^K\}$}
$\mathbf{\Sigma}_x = \mX \mX^T$, $\mathbf{\Sigma}_y = \mY \mY^T$, $\mathbf{\Sigma}_{xy} = \mX \mY^T$ \\
\For{$k \in {1 \dots K}$}{
$\vu^k, \vv^k \leftarrow GCCA_{\Theta}(\mathbf{\Sigma}_x,\mathbf{\Sigma}_y, \mathbf{\Sigma}_{xy}, \mL_1, \mL_2 )$
\\
$ \mathbf{\Sigma}_{xy} \leftarrow \mathbf{\Sigma}_{xy} - \dfrac{\langle \mathbf{\Sigma}_{xy} , \vu^k (\vv^k)^T \rangle}{\norm{\vu^k (\vv^k)^T}_2}.\vu^k (\vv^k)^T$ \\
$ \mathbf{\Sigma}_{xy} \leftarrow \mathbf{\Sigma}_{xy}/ \norm{\mathbf{\Sigma}_{xy}}_F$
}
\caption{{$K$-Graph-based SCCA (K-$GCCA$)} }
\label{algo:gscca-k}
\end{algorithm}
\subsection{Feature Embedding Module}
\label{feat_embed}
We propose a feature embedding module that utilizes the $K$-GCCA algorithm. The module generates embedding matrices $\mU$ and $\mV$, from which embeddings $\mX_e \in \R^{K \times n}, \mY_e  \in \R^{K \times n}$ of data matrices $\mX$ and $\mY$ are generated as linear combinations $\mX_e = \mU^T \mX, \mY_e = \mV^T \mY$ and the final embedding of the two modalities is obtained as $
   \mZ_e = \begin{bmatrix}
        \mX_e \\
        \mY_e 
     \end{bmatrix}~\in~\R^{2K \times n}.
$
\section{Experiments and Results}
\label{sec:experiments}

We first compare the 1-GCCA method with 1-SCCA on simulated data and 
breast cancer data to show that 1-GCCA learns better correlations than 1-SCCA.
Next, we present correlations resulting from the multi-dimensional extensions, with $K=100$. Finally, we employ the 100-GCCA embedding module to embed breast cancer data for one year survival prediction. 
All code and data is made available\footnote{\url{https://github.com/svaishnavi411/cca_fusion}}.

In our experiments, we consider two different ways of defining the underlying graph structures in the algorithm: (i) using the squared correlation between features within the same modality as edge weights, 
and (ii) directly from prior knowledge. 

\subsection{Correlations on Simulated Data (1-GCCA)}

Following previous works~\cite{du2015gn,chen2012efficient}, we generate data as follows. 
To construct $\vu$, we generate a random fully-connected graph with all edge-weights as 1. 
The eigenvectors corresponding to the first $l$ {non-zero} eigenvalues of the corresponding graph Laplacian are combined using randomly generated weights, and normalized to unit-norm to generate $\vu$. 
The vector $\vv$ is set to be \{10 3s, 10 -1.5s, 10 1s, 10 2s, 60 0s\} 
and normalized to unit-norm. 
The data matrices $\mX$ and $\mY$ are then generated such that columns $\mX_i \sim \calN(\vu w_i, \sigma^2 \mI_{p\times p}$) and $\mY_i \sim \calN(\vv w_i, \sigma^2 \mathbf{\Sigma}_v$), where $\mathbf{\Sigma}_v[i, j] = \exp(-|v_i - v_j|)$ and $w_i \sim \calN(0, 1)$. 
Data is generated for $n=1000$ samples with $p=q=100$, $l\in \{5, 10, 25, 50\}$ and $\sigma \in \{0.5, 0.75\}$, $25$ times for each setting. Across settings and repetitions, $\vv$ is fixed, while $\vu$ and $\mL_1$ change. For each setting and each repetition, the simulated data is 
and split into 50\%-10\%-40\% train, validation and test sets. 

We compare 1-SCCA and 1-GCCA, with the best hyperparameters chosen on the validation set. For 1-GCCA, we consider two methods of computing $\mL_1$: (i) from samples (1-GCCA), and (ii) feeding the true underlying $\mL$ as prior knowledge (1-GCCA-Prior).
Table~\ref{table:simulations} shows the mean and standard deviation of different error metrics on the test set across different parameters ($l\in \{5, 10, 25, 50\}$ and $\sigma \in \{0.5, 0.75\}$). 1-GCCA-Prior has the advantage of the true graph and outperforms 1-SCCA and 1-GCCA across all error metrics. 1-GCCA performs better than 1-SCCA in the estimation of $\vv$ and correlation $\rho$. 

\begin{table}[h]
\centering
\caption{Simulated data: Overall mean and standard deviations of absolute cosine distance $d_{\cos}$ in estimation of vectors $\vu$, $\vv$, absolute error in correlation $\rho$ and relative spectral frequency $\vu^T \mL_1 \vu$. Lower values desired.}
\resizebox{0.48\textwidth}{!}{%
\label{table:simulations}
\begin{tabular}{|c|c|cc|}
\hline
Error & 1-SCCA & 1-GCCA & 1-GCCA-Prior \\
\hline
$d_{\cos}(\vu, \hat\vu)$ &	44.54 $\pm$ 34.26 & 	44.24 $\pm$ 24.31 & 	31.95 $\pm$ 20.64 \\
$d_{\cos}(\vv, \hat\vv)$ & 	43.89 $\pm$ 41.42 & 	15.83 $\pm$ 20.61 & 	11.86 $\pm$ 16.17 \\
$|\rho - \hat\rho|$ & 	13.56 $\pm$ 13.37 & 	10.19 $\pm$ 8.61 & 	6.30 $\pm$ 6.75 \\
$\dfrac{|\vu^T \mL_1 \vu - \hat\vu^T \mL_1 \hat\vu|}{|\vu^T \mL_1 \vu|}$ & 124.24 $\pm$ 142.92 & 	105.74 $\pm$ 105.81 & 	42.32 $\pm$ 60.02 \\
\hline
\end{tabular}}
\end{table}

\subsection{Correlations on Breast Cancer Data (1-GCCA)}

We work on histology imaging and gene expressions from the TCGA breast adenocarcinoma (BRCA) dataset of $n=974$ patients to demonstrate the potential on real data. 

The imaging data was acquired from the National Cancer Institute's
\href{https://portal.gdc.cancer.gov/}{Genomic Data Commons portal}. For the histology images, we downloaded the nuclei segmentations from a recently published adversarial learning framework~\cite{hou2019robust}. The nuclei segmentations are provided each patient in patches. We randomly selected $25$ patches of size 2000 x 2000 pixels for each patient and fed the histology patch and segmentation mask to the CellProfiler tool to extract area, shape and texture properties for each nuclei and cell in the patch. Averaging these features across different patches yielded $213$-dimensional imaging feature vectors for each patient. 

The gene expression data was downloaded from the \href{http://firebrowse.org/}{FireBrowse platform}. We evaluated the most variant genes using the coefficient of variation ($\sigma/\mu$) of the log2-transformed expression values. We selected the top $500, 800, 1000$ and $3000$ genes and the corresponding z-scores of the genes serve as the genomic feature vector for each patient. 
To extract prior-knowledge dependencies between genes we used the protein-protein interactions from the \href{https://string-db.org/}{STRING database} which captures biologically meaningful physical and functional interactions between proteins and assigned weights between any two interacting proteins' corresponding genes.

\begin{table}[t]
\centering
\caption{TCGA-BRCA: Mean and standard deviations of correlation coefficients on test set across folds. $p$ denotes number of genes used in the correlation analysis. } 
\label{table:brca_results}
\resizebox{0.45\textwidth}{!}{
\begin{tabular}{|c|c|cc|}
\hline
{$p$} & {SCCA} & {1-GCCA} & {1-GCCA-Prior} \\
\hline
$500$  & 0.39 $\pm$ 0.11	 & 0.50 $\pm$ 0.09	 & 0.51 $\pm$ 0.08 \\
$800$ &  0.41 $\pm$ 0.11	 & 0.55 $\pm$ 0.09	 & 0.52 $\pm$ 0.06  \\
$1000$ & 0.46 $\pm$ 0.02	 & 0.55 $\pm$ 0.09	 & 0.49 $\pm$ 0.04 \\
$3000$ & 0.36 $\pm$ 0.18	 & 0.56 $\pm$ 0.09	 & 0.40 $\pm$ 0.13  \\
\hline
\end{tabular} 
}
\end{table}

\begin{table}[t]
\centering
\caption{TCGA-BRCA: Mean and standard deviations of the sum of correlations across the first 100 variates. $p$ denotes number of genes used in the correlation analysis.}
\label{table:brca_results_100}
\resizebox{0.45\textwidth}{!}{
\begin{tabular}{|c|c|cc|}
\hline
{$p$} & {K-SCCA} & {K-GCCA} & {K-GCCA-Prior} \\
\hline
500	 & 19.75 $\pm$ 3.66	 & 9.01 $\pm$ 0.81	 & 8.84 $\pm$ 0.44 \\
800	 & 20.52 $\pm$ 5.49	 & 9.42 $\pm$ 2.47	 & 9.59 $\pm$ 3.01 \\
1000	 & 20.79 $\pm$ 5.87	 & 11.66 $\pm$ 1.78	 & 10.97 $\pm$ 1.45 \\
3000	 & 25.56 $\pm$ 3.73	 & 11.85 $\pm$ 3.60	 & 12.65 $\pm$ 1.62 \\
\hline
\end{tabular} 
}
\end{table} 
We evaluate the methods with 5-fold cross validation using 544-137-292 patients in training-validation-test sets respectively. Table~\ref{table:brca_results} reports results on the test set across folds. It is observed that 1-GCCA and 1-GCCA-Prior reveal higher correlations across different number of most variant genes ($p$). Further, as number of genes $p$ increases, 1-GCCA is able to learn higher correlated embeddings.

\subsection{Correlations on Breast Cancer Data (K-GCCA)}

Next, we evaluate the K-GCCA proposed in Algorithm~\ref{algo:gscca-k}. Employing a similar approach with respect to SCCA, we obtain  K-SCCA. The sum of the correlation coefficients of the first 100 directions $\{\vu_1 \dots \vu_{100}\}$ and $\{\vv_1 \dots \vv_{100}\}$ are reported in Table~\ref{table:brca_results_100}. It is observed that K-SCCA yields higher sums across number of genes. Increasing the number of genes considered improves the sum of correlations for all methods.

\subsection{Survival Prediction on Breast Cancer (K-GCCA)}

\begin{table}[t]
\centering
\caption{TCGA-BRCA: Mean and standard deviations of different metrics on F1 \% scores of one year survival prediction problem using single modalities, early fusion, late fusion and CCA-based fusion modules.}
\label{table:survival_classification}
\centering
\resizebox{0.48\textwidth}{!}{%
\begin{tabular}{|l|cccc|}
\hline
Method 	& 500 & 800 & 1000 & 3000 \\
\hline
Genomics	 & 55.44 $\pm$ 1.90	 & 58.39 $\pm$ 2.56	 & 54.85 $\pm$ 2.80	 & 58.36 $\pm$ 2.29 \\
Imaging		 & 60.92 $\pm$ 1.17	 & 60.92 $\pm$ 1.17	 & 60.92 $\pm$ 1.17	 & 60.92 $\pm$ 1.17 \\
Early Fusion	 & 57.06 $\pm$ 5.55	 & 58.61 $\pm$ 3.53	 & 58.98 $\pm$ 1.01	 & 60.97 $\pm$ 1.75 \\
Late Fusion	 & 53.44 $\pm$ 2.19	 & 53.80 $\pm$ 3.20	 & 52.02 $\pm$ 3.73	 & 53.64 $\pm$ 4.04 \\
\hline
100-SCCA	 & 57.52 $\pm$ 2.91	 & 59.09 $\pm$ 3.27	 & 58.23 $\pm$ 2.57	 & 56.53 $\pm$ 4.53 \\
100-GCCA	 & 56.36 $\pm$ 3.16	 & 57.11 $\pm$ 3.02	 & 57.92 $\pm$ 0.97	 & 58.69 $\pm$ 2.16 \\
100-GCCA-P	 & 56.23 $\pm$ 2.23	 & 58.52 $\pm$ 4.75	 & 57.42 $\pm$ 1.84	 & 57.71 $\pm$ 2.45 \\
\hline
\end{tabular}
}
\end{table}
We make use of the proposed $K$-GCCA fusion module with $K=100$ to generate the embedding $\mZ_e$ to predict one-year survival of the TCGA-BRCA patients as a binary classification problem. 
We feed $\mZ_e$ to a random forest of $100$ estimators with maximum depth $d=50$. For baselines, we compare these features to $\mX$ only (Genomics), $\mY$ only (Imaging) and $[\mX^T, \mY^T]^T$ (Early fusion). Further, we combine the predictions of genomics and imaging in a simple late fusion module (Late fusion). 
We further utilize the embeddings from 100-SCCA as features for the random forest (100-SCCA). 
As earlier, we compare the construction of graphs from data (100-GCCA) and those from prior knowledge (100-GCCA-P).

The accuracy, support-weighted F1 and support-weighted AUC scores for the same test set as before are reported in Table~\ref{table:survival_classification}. Among all CCA-fusion methods, we observe that 100-SCCA works best for the lower number of genes ($p=\{500, 800, 1000\}$), while 100-GCCA and 100-GCCA-P work best for the largest number of genes ($p=3000$). 
\section{Conclusion}
\label{sec:conclusions}
In this work, we proposed a novel feature embedding module for multi-modality fusion with two modalities which generates well-correlated low-dimensional embeddings by taking into account intra-modality correlations. 
We first demonstrated the importance of accounting for intra-modality correlations in the CCA formulation. We showed that our proposed feature embedding module generates low-dimensional embeddings of the two modalities while preserving the information important for one-year survival prediction of breast cancer patients. In the future we will investigate the use of better deflation schemes for generating higher-dimensional embeddings, and conduct an extensive study across different cancers to comprehensively evaluate CCA-based fusion methods for cancer survival prediction.
\newpage
\section{Compliance with Ethical Standards}
\label{sec:ethics}
This research study was conducted retrospectively using
human subject data made available in open access by TCGA Research Network: \url{ https://www.cancer.gov/tcga}. Ethical approval was not required as confirmed by
the license attached with the open access data.
\section{Acknowledgments}
\label{sec:acknowledgments}
This project has been funded by the Jump ARCHES endowment through the Health Care Engineering Systems Center and the IBM-Illinois C3SR center.

\bibliographystyle{IEEEbib}
\bibliography{references}
\end{document}